\pdfoutput=1
\documentclass[11pt,a4paper]{article}
\usepackage[hyperref]{acl2019nopatch}
\usepackage{times}
\usepackage{latexsym}
\usepackage{microtype}
\usepackage{amsmath}
\usepackage{amsfonts}
\usepackage{bibentry}
\usepackage{booktabs}
\usepackage{multirow}
\usepackage{tabularx}
\usepackage{xcolor}
\usepackage{url}
\usepackage{enumerate}
\usepackage{subfigure}
\usepackage{enumitem}
\usepackage{multicol}
\usepackage[]{linguex}
\usepackage{adjustbox}

\usepackage{cjhebrew}

\usepackage{tikz}
\usetikzlibrary{bayesnet}
\usepackage{tikz-dependency}
\usetikzlibrary{shapes.arrows}
\usepackage{pgfplots}

\usepackage{cleveref}
\crefname{section}{\S}{\S\S}
\Crefname{section}{\S}{\S\S}
\crefname{table}{Tab.}{}
\crefname{figure}{Fig.}{}
\crefname{algorithm}{Algorithm}{}
\crefname{equation}{eq.}{}
\crefname{appendix}{App.}{}
\crefname{ExNo}{Sentence}{}
\crefformat{section}{\S#2#1#3}  % remove space between section symbol and the number

\newcolumntype{C}{>{\centering\arraybackslash}X}

\newcommand{\vm}{\mathbf{m}}
\newcommand{\vp}{\mathbf{p}}

\newcommand{\word}[1]{\textit{#1}}

\newcommand{\Plm}{P_{\textit{lm}}}
\newcommand{\psif}{\textcolor{black!40!red!30!blue}{\psi}}
\newcommand{\phif}{\textcolor{black!60!green}{\phi_i}}

\newcommand{\overbar}[1]{\mkern 1.5mu\overline{\mkern-1.5mu#1\mkern-1.5mu}\mkern 1.5mu}

\makeatletter
\def\@selfnt{}
\makeatother

\def\aclpaperid{529} %  Enter the acl Paper ID here

\title{Counterfactual Data Augmentation for Mitigating Gender Stereotypes in Languages with Rich Morphology}
\author{Ran Zmigrod$^1$ \hspace{1.5em} Sabrina J. Mielke$^2$ \hspace{1.5em} Hanna Wallach$^3$ \hspace{1.5em} Ryan Cotterell$^1$ \\
  ${}^1$ University of Cambridge \hspace{1.5em} ${}^2$ Johns Hopkins University \hspace{1.5em} ${}^3$ Microsoft Research \\
  {\tt rz279@cam.ac.uk} \;\; {\tt sjmielke@jhu.edu} \\ {\tt wallach@microsoft.com} \;\; {\tt rdc42@cam.ac.uk} \\}

\date{}

\aclfinaltrue

\pgfplotsset{compat=1.11,
    /pgfplots/ybar legend/.style={
    /pgfplots/legend image code/.code={%
       \draw[##1,/tikz/.cd,yshift=-0.25em]
        (0cm,0cm) rectangle (3pt,0.8em);},
   },
}

\begin{document}
\maketitle

\thispagestyle{plain}
\pagestyle{plain}

\begin{abstract}
Gender stereotypes are manifest in most of the world's languages and
are consequently propagated or amplified by NLP systems. Although
research has focused on mitigating gender stereotypes in English, the
approaches that are commonly employed produce ungrammatical sentences
in morphologically rich languages. We present a novel approach for
converting between masculine-inflected and feminine-inflected
sentences in such languages. For Spanish and Hebrew, our approach
achieves $F_1$ scores of 82\% and 73\% at the level of tags and
accuracies of 90\% and 87\% at the level of forms. By evaluating our
approach using four different languages, we show that, on average, it
reduces gender stereotyping by a factor of 2.5 without any sacrifice
to grammaticality.
\end{abstract}

\section{Introduction}

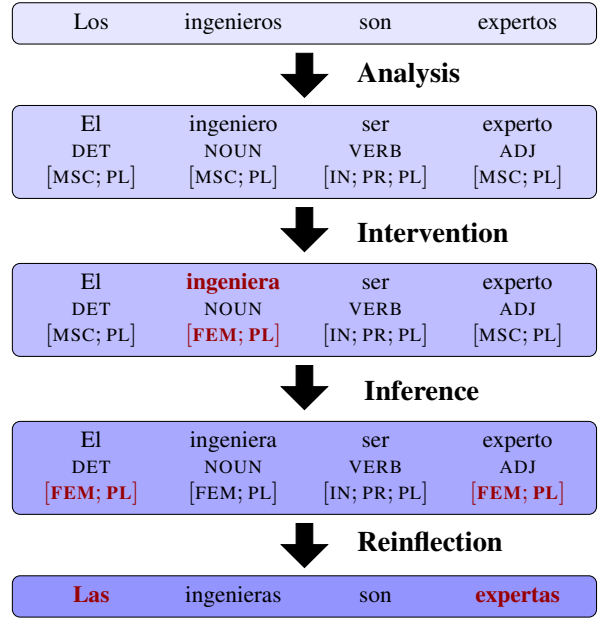
\begin{figure}[t!]
    \centering
    \small
    \begin{adjustbox}{width=\columnwidth}
    \begin{tikzpicture}
        \node[rounded corners=3pt, draw, fill=blue!10]{
            \begin{tabularx}{0.48\textwidth}{CCCC}
            Los & ingenieros & son & expertos
            \end{tabularx}};
        \node[draw, single arrow,
            minimum height=6mm, minimum width=4mm,
            single arrow head extend=2mm, fill=black,
            anchor=west, rotate=-90] at (0,-0.4) {};
        \node[right] at (0.6,-0.7) {\normalsize \textbf{Analysis}};
        \node[rounded corners=3pt, draw, fill=blue!18] at (0, -1.75) {
            \begin{tabularx}{0.48\textwidth}{CCCC}
            El & ingeniero & ser & experto \\
            $\textsc{det}$ & $\textsc{noun}$ & $\textsc{verb}$ & $\textsc{adj}$ \\
            $[\textsc{msc};\textsc{pl}]$ & $[\textsc{msc};\textsc{pl}]$ & $[\textsc{in};\textsc{pr};\textsc{pl}]$ & $[\textsc{msc};\textsc{pl}]$
            \end{tabularx}};
        \node[draw, single arrow,
            minimum height=6mm, minimum width=4mm,
            single arrow head extend=2mm, fill=black,
            anchor=west, rotate=-90] at (0,-2.5) {};
        \node[right] at (0.6,-2.85) {\normalsize \textbf{Intervention}};
        \node[rounded corners=3pt, draw, fill=blue!26] at (0, -3.9) {
            \begin{tabularx}{0.48\textwidth}{CCCC}
            El & \textcolor{black!40!red}{\textbf{ingeniera}} & ser & experto \\
            $\textsc{det}$ & $\textsc{noun}$ & $\textsc{verb}$ & $\textsc{adj}$ \\
            $[\textsc{msc};\textsc{pl}]$ & \textcolor{black!40!red}{\textbf{$[\textsc{fem};\textsc{pl}]$}} & $[\textsc{in};\textsc{pr};\textsc{pl}]$ & $[\textsc{msc};\textsc{pl}]$
            \end{tabularx}};
        \node[draw, single arrow,
            minimum height=6mm, minimum width=4mm,
            single arrow head extend=2mm, fill=black,
            anchor=west, rotate=-90] at (0,-4.65) {};
        \node[right] at (0.6,-5) {\normalsize \textbf{ Inference}};
        \node[rounded corners=3pt, draw, fill=blue!34] at (0, -6.05) {
            \begin{tabularx}{0.48\textwidth}{CCCC}
            El & ingeniera & ser & experto \\
            $\textsc{det}$ & $\textsc{noun}$ & $\textsc{verb}$ & $\textsc{adj}$ \\
            \textcolor{black!40!red}{\bf $[\textsc{fem};\textsc{pl}]$} & $[\textsc{fem};\textsc{pl}]$ &
            $[\textsc{in};\textsc{pr};\textsc{pl}]$ & \textcolor{black!40!red}{\bf $[\textsc{fem};\textsc{pl}]$}
            \end{tabularx}};
        \node[draw, single arrow,
            minimum height=6mm, minimum width=4mm,
            single arrow head extend=2mm, fill=black,
            anchor=west, rotate=-90] at (0,-6.8) {};
            \node[right] at (0.6,-7.05) {\normalsize \textbf{Reinflection}};
        \node[rounded corners=3pt, draw, fill=blue!42] at (0, -7.8) {
            \begin{tabularx}{0.48\textwidth}{CCCC}
            \textcolor{black!40!red}{\bf Las} & ingenieras & son & \textcolor{black!40!red}{\bf expertas}
            \end{tabularx}};
    \end{tikzpicture}
    \end{adjustbox}
    \caption{Transformation of \word{Los ingenieros son expertos}
      (i.e., \word{The male engineers are skilled}) to \word{Las
        ingenieras son expertas} (i.e., \word{The female engineers are
        skilled}). We extract the properties of each word in the
      sentence. We then fix a noun and its tags and infer the manner
      in which the remaining tags must be updated. Finally, we
      reinflect the lemmata to their new forms.\looseness=-1}
    \label{fig:pipeline}
\end{figure}

One of the biggest challenges faced by modern natural language
processing (NLP) systems is the inadvertent replication or
amplification of societal biases. This is because NLP systems depend
on language corpora, which are inherently ``not objective; they are
creations of human design'' \cite{crawford2013}. One type of societal
bias that has received considerable attention from the NLP community
is gender stereotyping~\cite{garg2018, W17-1609, sutton2018}. Gender
stereotypes can manifest in language in overt ways. For example, the
sentence \word{he is an engineer} is more likely to appear in a corpus
than \word{she is an engineer} due to the current gender disparity in
engineering. Consequently, any NLP system that is trained such a
corpus will likely learn to associate \word{engineer} with men, but
not with women~\cite{DBLP:conf/fat/De-ArteagaRWCBC19}.

To date, the NLP community has focused primarily on approaches for
detecting and mitigating gender stereotypes in
English~\cite{bolukbasi2016, dixon2018, zhao2017}. Yet, gender
stereotypes also exist in other languages because they are a function
of society, not of grammar. Moreover, because English does not mark
grammatical gender, approaches developed for English are not
transferable to morphologically rich languages that exhibit gender
agreement \cite{corbett1991}. In these languages, the words in a
sentence are marked with morphological endings that reflect the
grammatical gender of the surrounding nouns. This means that if the
gender of one word changes, the others have to be updated to match. As
a result, simple heuristics, such as augmenting a corpus with
additional sentences in which \word{he} and \word{she} have been
swapped~\cite{zhao2018gender}, will yield ungrammatical
sentences. Consider the Spanish phrase \word{el ingeniero experto}
(\word{the skilled engineer}). Replacing \word{ingeniero} with
\word{ingeniera} is insufficient---\word{el} must also be replaced
with \word{la} and \word{experto} with \word{experta}.

\begin{figure*}[t]
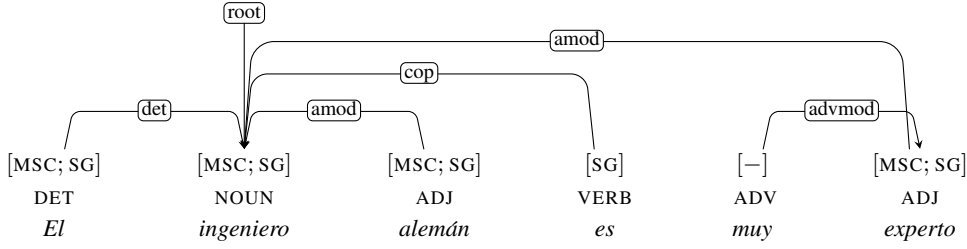

    \centering
    \small
    \begin{dependency}
    \begin{deptext}[column sep=1.1cm, row sep=.3ex]
    $[\textsc{msc};\textsc{sg}]$ \&
    $[\textsc{msc};\textsc{sg}]$ \&
    $[\textsc{msc};\textsc{sg}]$ \&
    $[\textsc{sg}]$ \&
    $[-]$ \&
    $[\textsc{msc};\textsc{sg}]$ \\
    $\textsc{det}$ \& $\textsc{noun}$ \& $\textsc{adj}$ \& $\textsc{verb}$ \& \ $\textsc{adv}$ \& $\textsc{adj}$ \\
    \word{El} \& \word{ingeniero} \& \word{alem\'{a}n} \& \word{es} \& \word{muy} \& \word{experto} \\
    \end{deptext}
    \depedge[style={font=\normalsize}]{1}{2}{det}
    \deproot[edge unit distance=3.5ex, style={font=\normalsize}]{2}{root}
    \depedge[style={font=\normalsize}]{3}{2}{amod}
    \depedge[style={font=\normalsize}]{4}{2}{cop}
    \depedge[edge unit distance=2.5ex, style={font=\normalsize}]{6}{2}{amod}
    \depedge[style={font=\normalsize}]{5}{6}{advmod}
    \end{dependency}
    \caption{Dependency tree for the sentence \word{El ingeniero alem\'{a}n es muy experto}.}
    \label{fig:tree}
\end{figure*}

In this paper, we present a new approach to counterfactual data
augmentation \citep[CDA;][]{DBLP:journals/corr/abs-1807-11714} for
mitigating gender stereotypes associated with animate\footnote{Specifically, we consider a noun to be animate if WordNet considers \word{person} to be a hypernym of that noun.} nouns (i.e.,
nouns that represent people) for morphologically rich languages. We
introduce a Markov random field with an optional neural
parameterization that infers the manner in which a sentence must
change when altering the grammatical gender of particular nouns. We
use this model as part of a four-step process, depicted in
\cref{fig:pipeline}, to reinflect entire sentences following an
intervention on the grammatical gender of one word. We intrinsically
evaluate our approach using Spanish and Hebrew, achieving tag-level
$F_1$ scores of 83\% and 72\% and form-level accuracies of 90\% and
87\%, respectively. We also conduct an extrinsic evaluation using four
languages. Following \newcite{DBLP:journals/corr/abs-1807-11714}, we
show that, on average, our approach reduces gender stereotyping in
neural language models by a factor of 2.5 without sacrificing
grammaticality.

\section{Gender Stereotypes in Text}
\label{sec:gender}

Men and women are mentioned at different rates in
text~\cite{coates2015women}. This problem is exacerbated in certain
contexts. For example, the sentence \word{he is an engineer} is more
likely to appear in a corpus than \word{she is an engineer} due to the
current gender disparity in engineering. This imbalance in
representation can have a dramatic downstream effect on NLP systems
trained on such a corpus, such as giving preference to male engineers
over female engineers in an automated resum\'{e} filtering
system. Gender stereotypes of this sort have been observed in word
embeddings \cite{bolukbasi2016,sutton2018}, contextual word embeddings
\cite{zhao2019}, and co-reference resolution systems \cite{N18-2002,
  zhao2018gender} \textit{inter alia}.

\paragraph{A quick fix: swapping gendered words.}
One approach to mitigating such gender stereotypes is counterfactual
data augmentation
\citep[CDA;][]{DBLP:journals/corr/abs-1807-11714}. In English, this
involves augmenting a corpus with additional sentences in which
gendered words, such as \word{he} and \word{she}, have been swapped to
yield a balanced representation. Indeed, \newcite{zhao2018gender}
showed that this simple heuristic significantly reduces gender
stereotyping in neural co-reference resolution systems, without harming
system performance. Unfortunately, this approach is only applicable to
English and other languages with little morphological inflection. When
applied to morphologically rich languages that exhibit gender
agreement, it yields ungrammatical sentences.

\paragraph{The problem: inflected languages.}

Many languages, including Spanish and Hebrew, have gender inflections
for nouns, verbs, and adjectives---i.e., the words in a sentence are
marked with morphological endings that reflect the grammatical gender
of the surrounding nouns.\footnote{The number of grammatical genders
  varies for different languages, with two being the most common
  non-zero number \citep{wals}. The languages that we use in our
  evaluation have two grammatical genders (male, female).} This means
that if the gender of one word changes, the others have to be updated
to preserve morpho-syntactic agreement~\cite{features}. Consider the
following example from Spanish, where we wish to transform
\cref{sent:msc} to \cref{sent:fem}. (Parts of words that mark gender
are depicted in bold.) This task is not as simple as replacing
\word{el} with \word{la}---\word{ingeniero} and \word{experto} must
also be reinflected. Moreover, the changes required for one language
are not the same as those required for another (e.g., verbs are marked
with gender in Hebrew, but not in Spanish).

{\small
\exg. \word{\textbf{El}} \word{ingenier\textbf{o}} \word{alem\'{a}n} \word{es} \word{muy} \word{expert\textbf{o}.}\\
The.\textsc{msc}.\textsc{sg} engineer.\textsc{msc}.\textsc{sg} German.\textsc{msc}.\textsc{sg} is.\textsc{in}.\textsc{pr}.\textsc{sg} very skilled.\textsc{msc}.\textsc{sg}\\
\textcolor{gray}{\mbox{}\\(The German engineer is very skilled.)} \label{sent:msc}

\mbox{}\\[-4em]

\exg. \word{\textbf{La}} \word{ingenier\textbf{a}} \word{aleman\textbf{a}} \word{es} \word{muy} \word{expert\textbf{a}.}\\
The.\textsc{fem}.\textsc{sg} engineer.\textsc{fem}.\textsc{sg} German.\textsc{fem}.\textsc{sg} is.\textsc{in}.\textsc{pr}.\textsc{sg} very skilled.\textsc{fem}.\textsc{sg}\\
\textcolor{gray}{\mbox{}\\(The German engineer is very skilled.)} \label{sent:fem}

}

\paragraph{Our approach.}\label{par:cda}

Our goal is to transform sentences like \cref{sent:msc} to
\cref{sent:fem} and vice versa. To the best of our knowledge, this
task has not been studied previously. Indeed, there is no existing
annotated corpus of paired sentences that could be used to train a
supervised model. As a result, we take an
unsupervised\footnote{Because we do not have any direct supervision
  for the task of interest, we refer to our approach as being
  unsupervised even though it does rely on annotated linguistic
  resources.}  approach using dependency trees, lemmata,
part-of-speech (POS) tags, and morpho-syntactic tags from Universal
Dependencies corpora \citep[UD;][]{ud}. Specifically, we propose the
following process:\looseness=-1

\begin{enumerate}[topsep=0pt,itemsep=-1ex,partopsep=1ex,parsep=1ex]
\itemsep0em
\item Analyze the sentence (including parsing, morphological analysis,
  and lemmatization).
\item Intervene on a gendered word.
\item Infer the new morpho-syntactic tags.
\item Reinflect the lemmata to their new forms.
\end{enumerate}
This process is depicted in \cref{fig:pipeline}.  The primary
technical contribution is a novel Markov random field for performing
step 3, described in the next section.\looseness=-1

\section{A Markov Random Field for Morpho-Syntactic Agreement}\label{sec:mrf}

In this section, we present a Markov random field \citep[MRF;][]{mrf}
for morpho-syntactic agreement. This model defines a joint
distribution over sequences of morpho-syntactic tags, conditioned on a
labeled dependency tree with associated part-of-speech tags. Given an
intervention on a gendered word, we can use this model to infer the
manner in which the remaining tags must be updated to preserve
morpho-syntactic agreement.\looseness=-1

A dependency tree for a sentence (see \cref{fig:tree} for an example)
is a set of ordered triples $(i, j, \ell)$, where $i$ and $j$ are
positions in the sentence (or a distinguished root symbol) and $\ell
\in L$ is the label of the edge $i \rightarrow j$ in the tree; each
position occurs exactly once as the first element in a triple. Each
dependency tree $T$ is associated with a sequence of morpho-syntactic
tags $\vm = m_1, \ldots, m_{|T|}$ and a sequence of part-of-speech
(POS) tags $\vp = p_1, \ldots, p_{|T|}$. For example, the tags $m \in
M$ and $p \in P$ for \word{ingeniero} are $[\textsc{msc};
  \textsc{sg}]$ and $\textsc{noun}$, respectively, because
\word{ingeniero} is a masculine, singular noun. For notational
simplicity, we define ${\cal M} = M^{|T|}$ to be the set of all
length-$|T|$ sequences of morpho-syntactic tags.

\begin{figure*}[t!]
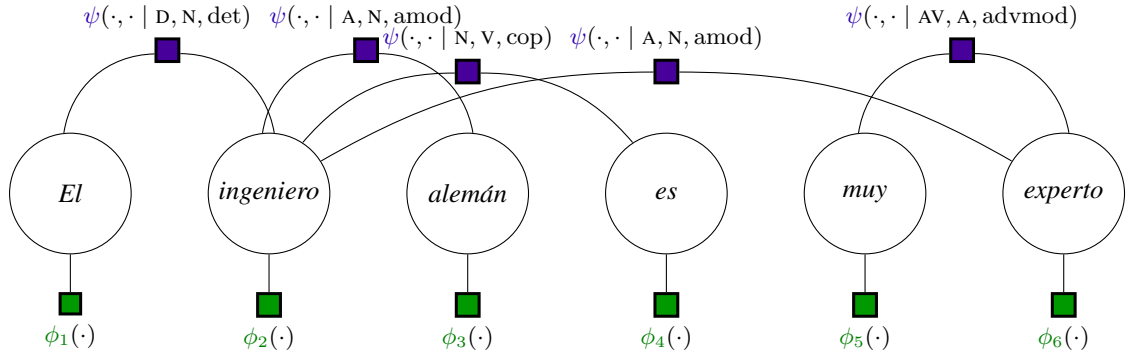

    \centering
    \small
    \tikz{
        % Nodes
        \node[latent, minimum size=1.6cm, label=below:$\ \ \ \ \ \ $ ]
        (1) {\word{El}}; %
        \node[latent, minimum size=1.6cm, right=1cm of 1, label=below:$\ \ \ \ \ \ \ \ \ \ \ \ \ \ \ \ \ \ \ \ \ $ ]
        (2) {\word{ingeniero}}; %
        \node[latent, minimum size=1.6cm, right=1cm of 2, label=below:$\ \ \ \ \ \ \ \ \ \ \ \ \ \ \ \ $]
        (3) {\word{alem{\'a}n}}; %
        \node[latent, minimum size=1.6cm, right=1cm of 3, label=below:$\ \ \ \ \ \ $ ]
        (4) {\word{es}}; %
        \node[latent, minimum size=1.6cm, right=1cm of 4, label=below:$\ \ \ \ \ \ \ \ \ \ \ $ ]
        (5) {\word{muy}}; %
        \node[latent, minimum size=1.6cm, right=1cm of 5, label=below:$\ \ \ \ \ \ \ \ \ \ \ \ \ \ \ $ ]
        (6) {\word{experto}}; %
        % Unary factors
        \factor[below=0.5cm of 1, minimum size=0.27cm, fill=black!40!green, line width = 0.4mm, draw=black]  {f1}
            {below:$\textcolor{black!50!green}{\phi_1}(\cdot)$} {} {}%
        \factor[below=0.5cm of 2, minimum size=0.3cm, fill=black!40!green, line width = 0.4mm, draw=black]  {f2}
            {below:$\textcolor{black!50!green}{\phi_2}(\cdot)$} {} {}%
        \factor[below=0.5cm of 3, minimum size=0.3cm, fill=black!40!green, line width = 0.4mm, draw=black]  {f3}
            {below:$\textcolor{black!50!green}{\phi_3}(\cdot)$} {} {}%
        \factor[below=0.5cm of 4, minimum size=0.3cm, fill=black!40!green, line width = 0.4mm, draw=black]  {f4}
            {below:$\textcolor{black!50!green}{\phi_4}(\cdot)$} {} {}%
        \factor[below=0.5cm of 5, minimum size=0.3cm, fill=black!40!green, line width = 0.4mm, draw=black]  {f5}
            {below:$\textcolor{black!50!green}{\phi_5}(\cdot)$} {} {}%
        \factor[below=0.5cm of 6, minimum size=0.3cm, fill=black!40!green, line width = 0.4mm, draw=black]  {f6}
            {below:$\textcolor{black!50!green}{\phi_6}(\cdot)$} {} {}%
        % Binary factors
        \factor[right=0.3cm of 1, yshift=1.9cm, minimum size=0.3cm, fill=black!30!red!40!blue, line width = 0.4mm, draw=black]
            {f12} {$\psif(\cdot, \cdot \mid \textsc{d}, \textsc{n}, \mathrm{det})$} {} {}%
        \factor[right=0.3cm of 2, yshift=1.9cm, minimum size=0.3cm, fill=black!30!red!40!blue, line width = 0.4mm, draw=black]
            {f32} {$\psif(\cdot, \cdot \mid \textsc{a}, \textsc{n}, \mathrm{amod})$} {} {}%
        \factor[above=0.64cm of 3, minimum size=0.3cm, fill=black!30!red!40!blue, line width = 0.4mm, draw=black]
            {f42} {$\psif(\cdot, \cdot \mid \textsc{n}, \textsc{v}, \mathrm{cop})$} {} {}%
        \factor[right=0.3cm of 5, yshift=1.9cm, minimum size=0.3cm, fill=black!30!red!40!blue, line width = 0.4mm, draw=black]
            {f56} {$\psif(\cdot, \cdot \mid \textsc{av}, \textsc{a}, \mathrm{advmod})$} {} {}%
        \factor[above=0.64cm of 4, minimum size=0.3cm, fill=black!30!red!40!blue, line width = 0.4mm, draw=black]
            {f62} {$\psif(\cdot, \cdot \mid \textsc{a}, \textsc{n}, \mathrm{amod})$} {} {}%
        % Unary Connections
        \factoredge {1} {f1} {} ; %
        \factoredge {2} {f2} {} ; %
        \factoredge {3} {f3} {} ; %
        \factoredge {4} {f4} {} ; %
        \factoredge {5} {f5} {} ; %
        \factoredge {6} {f6} {} ; %
        % Binary Connections
        \factoredge[bend left=40] {1} {f12} {} ; %
        \factoredge[bend right=40] {2} {f12} {} ; %
        \factoredge[bend left=40] {2} {f32} {} ; %
        \factoredge[bend right=40] {3} {f32} {} ; %
        \factoredge[bend left=40] {5} {f56} {} ; %
        \factoredge[bend right=40] {6} {f56} {} ; %
        \factoredge[bend left=25] {2} {f42} {} ; %
        \factoredge[bend right=25] {4} {f42} {} ; %
        \factoredge[bend left=15] {2} {f62} {} ; %
        \factoredge[bend right=15] {6} {f62} {} ; %
    }
    \caption{Factor graph for the sentence \word{El ingeniero alem\'{a}n es muy experto}.}
    \label{fig:fg}
\end{figure*}

We define the probability of $\vm$ given $T$ and
$\vp$ as
\begin{align}
  &\text{Pr}(\vm \,|\, T, \vp) \propto {}\notag\\
  &\prod_{(i, j, \ell) \in T} \phif(m_i)\cdot\psif(m_i, m_j \,|\, p_i, p_j, \ell),
\label{eq:dist}
\end{align}
where the binary factor $\psif(\cdot, \cdot \,|\, \cdot, \cdot, \cdot)
\geq 0$ scores how well the morpho-syntactic tags $m_i$ and $m_j$
agree given the POS tags $p_i$ and $p_j$ and the label $\ell$. For
example, consider the $\mathrm{amod}$ (adjectival modifier) edge from
\word{experto} to \word{ingeniero} in \cref{fig:tree}. The factor
$\psif(m_i, m_j\,|\, \textsc{a}, \textsc{n}, \mathrm{amod})$ returns a
high score if the corresponding morpho-syntactic tags agree in gender
and number (e.g., $m_i=[\textsc{msc};\textsc{sg}]$ and
$m_j=[\textsc{msc};\textsc{sg}]$) and a low score if they do not
(e.g., $m_i=[\textsc{msc};\textsc{sg}]$ and
$m_j=[\textsc{fem};\textsc{pl}]$). The unary factor $\phif(\cdot) \geq
0$ scores a morpho-syntactic tag $m_i$ outside the context of the
dependency tree. As we explain in \cref{sec:constraint}, we use these
unary factors to force or disallow particular tags when
performing an intervention; we do not learn them. \Cref{eq:dist} is
normalized by the following partition function:\looseness=-1
\begin{align*}
  &Z(T, \vp) = {} \notag\\
&\sum_{\vm' \in {\cal M}}\prod_{(i, j, \ell) \in T} \phif(m'_i)\cdot\psif(m'_i, m'_j \mid  p_i, p_j, \ell).
\end{align*}
$Z(T, \vp)$ can be calculated using belief propagation; we provide the update equations that we use in \cref{sec:bp}.
Our model is depicted in \cref{fig:fg}. It is noteworthy that this model is
delexicalized---i.e., it considers only the labeled dependency tree
and the POS tags, not the actual words themselves.%\looseness=-1

\subsection{Parameterization}\label{sec:param}

We consider a linear parameterization and a neural parameterization of
the binary factor $\psif(\cdot, \cdot \,|\, \cdot, \cdot, \cdot)$.

\paragraph{Linear parameterization.}

We define a matrix $W(p_i, p_j, \ell)\in\mathbb{R}^{c\times c}$ for
each triple $(p_i, p_j, \ell)$, where $c$ is the number of
morpho-syntactic subtags. For example, $[\textsc{msc};\textsc{sg}]$
has two subtags $\textsc{msc}$ and $\textsc{sg}$. We then define
$\psif(\cdot, \cdot \,|\, \cdot, \cdot, \cdot)$ as follows:
\begin{align*}
\psif(m_i, m_j \mid p_i, p_j, \ell) = \exp{(\overbar{m}_i^{\top} W(p_i, p_j, \ell) \overbar{m}_j)}, \nonumber
\end{align*}
where $\overbar{m}_i \in \{0, 1\}^c$ is a multi-hot encoding of $m_i$.

\paragraph{Neural parameterization.}

As an alternative, we also define a neural parameterization of $W(p_i,
p_j, \ell)$ to allow parameter sharing among edges with different
parts of speech and labels:\looseness=-1
\begin{align*}
  &W(p_i, p_j, \ell) = \notag\\
  &\quad\exp{(U \tanh (V\,[\mathbf{e}(p_i); \mathbf{e}(p_j); \mathbf{e}(\ell)]))} \nonumber
\end{align*}
where $U \in\mathbb{R}^{c\times c\times n_1}$, $V \in\mathbb{R}^{n_1
  \times 3n_2}$, and $n_1$ and $n_2$ define the structure of the
neural parameterization and each $\mathbf{e}(\cdot) \in
\mathbb{R}^{n_2}$ is an embedding function.

\paragraph{Parameterization of $\phif$.}
\label{sec:constraint}

We use the unary factors only to force or disallow particular tags
when performing an intervention. Specifically, we define
\begin{equation}
\phif(m) = \begin{cases}
      \alpha & \text{if } m = m_i \\
      1 & \text{otherwise},
   \end{cases}
\end{equation}
where $\alpha >1$ is a strength parameter that determines the extent
to which $m_i$ should remain unchanged following an intervention. In
the limit as $\alpha \rightarrow \infty$, all tags will remain
unchanged except for the tag directly involved in the
intervention.\footnote{In practice, $\alpha$ is set using development data.}\looseness=-1

\subsection{Inference}\label{belief}

Because our MRF is acyclic and tree-shaped, we can use belief
propagation~\citep{pearl} to perform exact inference. The algorithm is
a generalization of the forward-backward algorithm for hidden Markov
models~ \cite{rabiner1986introduction}. Specifically, we pass messages
from the leaves to the root and vice versa. The marginal distribution
of a node is the point-wise product of all its incoming messages; the
partition function $Z(T, \vp)$ is the sum of any node's marginal
distribution.  Computing $Z(T, \vp)$ takes polynomial time
\cite{pearl}---specifically, ${\cal O}(n\cdot |M|^2)$ where $M$ is the
number of morpho-syntactic tags. Finally, inferring the
highest-probability morpho-syntactic tag sequence $\vm^{\star}$ given
$T$ and $\vp$ can be performed using the max-product modification to
belief propagation.

\subsection{Parameter Estimation}

We use gradient-based optimization. We treat the negative log-likelihood
${}-\log{(\text{Pr}(\vm \,|\, T, \vp))}$ as the loss function for tree
$T$ and compute its gradient using automatic
differentiation~\cite{DBLP:books/sp/Rall81}.
We learn the parameters of \cref{sec:param} by optimizing the negative log-likelihood using gradient descent.\looseness=-1

\section{Intervention}\label{sec:intervention}

As explained in \cref{sec:gender}, our goal is to transform sentences
like \cref{sent:msc} to \cref{sent:fem} by intervening on a gendered
word and then using our model to infer the manner in which the
remaining tags must be updated to preserve morpho-syntactic
agreement. For example, if we change the morpho-syntactic tag for
\word{ingeniero} from [\textsc{msc};\textsc{sg}] to
     [\textsc{fem};\textsc{sg}], then we must also update the tags for
     \word{el} and \word{experto}, but do not need to update the tag
     for \word{es}, which should remain unchanged as [\textsc{in};
       \textsc{pr}; \textsc{sg}]. If we intervene on the $i^\text{th}$
     word in a sentence, changing its tag from $m_i$ to $m_i'$, then
     using our model to infer the manner in which the remaining tags
     must be updated means using $\text{Pr}(\mathbf{m}_{-i} \,|\,
     m'_i, T, \vp)$ to identify high-probability tags for the
     remaining words.\looseness=-1

Crucially, we wish to change as little as possible when intervening on
a gendered word. The unary factors $\phif$ enable us to do exactly
this. As described in the previous section, the strength parameter
$\alpha$ determines the extent to which $m_i$ should remain unchanged
following an intervention---the larger the value, the less likely it
is that $m_i$ will be changed.\looseness=-1

\begin{table}[t]
    \centering
    \small
    \begin{tabular}{lclc} \toprule
Language & Accuracy & Language & Accuracy \\
\midrule
         French & 93.17 & Italian & 98.29 \\
         Hebrew & 95.16 & Spanish & 97.78 \\
         \bottomrule
    \end{tabular}
    \caption{Morphological reinflection accuracies.}
    \label{tab:reinflect}
\end{table}

Once the new tags have been inferred, the final step
is to reinflect the lemmata to their new forms. This task has received
considerable attention from the NLP
community~\citep{cotterell2016,cotterell-etal-2017-conll}. We use the inflection model
of \newcite{D18-1473}. This model conditions on the lemma $\mathbf{x}$
and morpho-syntactic tag $m$ to form a distribution over possible
inflections. For example, given \word{experto} and
$[\textsc{a};\textsc{fem};\textsc{pl}]$, the trained inflection model will
assign a high probability to \word{expertas}. We provide accuracies for the trained inflection model in \cref{tab:reinflect}.

\section{Experiments}

We used the Adam
optimizer \cite{DBLP:journals/corr/KingmaB14} to train both parameterizations of our model until the change in dev-loss was less than $10^{-5}$ bits.
We set $\beta=(0.9, 0.999)$ without tuning, and chose a learning rate of $0.005$ and weight decay factor of $0.0001$ after tuning.
We tuned $\log\alpha$ in the set $\{0.5, 0.75, 1, 2, 5, 10\}$ and chose $\log\alpha=1$.
For the neural
parameterization, we set $n_1 = 9$ and $n_2 = 3$ without any tuning. Finally, we
trained the inflection model using only gendered words.\looseness=-1

We evaluate our approach both intrinsically and extrinsically. For the
intrinsic evaluation, we focus on whether our approach yields the
correct morpho-syntactic tags and the correct reinflections. For the
extrinsic evaluation, we assess the extent to which using the resulting
transformed sentences reduces gender stereotyping in neural language
models.

\subsection{Intrinsic Evaluation}
\begin{table}
    \centering
    \small
    \begin{tabular}{ r c c } \toprule
        \textbf{Language} & \textbf{Training Size} & \textbf{Annotated Test Size} \\ \midrule
        Hebrew & 5,241 & 111 \\
        Spanish & 14,187 & 136 \\ \midrule
        French & 14,554 & -- \\
        Italian & 12,837 & -- \\
        % Italian & 12,837 & -- \\
        % Italian & 12,837 & -- \\
        \bottomrule
    \end{tabular}
    \caption{Language data.}
    \label{tab:data}
\end{table}

To the best of our knowledge, this task has not been studied
previously. As a result, there is no existing annotated corpus of paired
sentences that can be used as ``ground truth.'' We therefore annotated
Spanish and Hebrew sentences ourselves, with annotations made by native speakers of each language. Specifically, for each
language, we extracted sentences containing animate nouns from that
language's UD treebank. The average length of these extracted
sentences was 37 words. We then manually inspected each sentence,
intervening on the gender of the animate noun and reinflecting the
sentence accordingly. We chose Spanish and Hebrew because gender
agreement operates differently in each language. We provide corpus
statistics for both languages in the top two rows of \cref{tab:data}.\looseness=-1

We created a hard-coded $\psif(\cdot, \cdot \,|\, \cdot, \cdot,
\cdot)$ to serve as a baseline for each language. For Spanish, we only
activated values (i.e. set to numbers greater than zero) that relate adjectives and determiners to nouns; for
Hebrew, we only activated values that relate adjectives and verbs to
nouns. We created these language-specific baselines because gender agreement operates differently in each language.\looseness=-1

To evaluate our approach, we held all morpho-syntactic subtags fixed
except for gender. For each annotated sentence, we intervened on the
gender of the animate noun. We then used our model to infer which of
the remaining tags should be updated (updating a tag means swapping
the gender subtag because all morpho-syntactic subtags were held fixed
except for gender) and reinflected the lemmata. Finally, we used the
annotations to compute the tag-level $F_1$ score and the form-level
accuracy, excluding the animate nouns on which we intervened.\looseness=-1

\begin{table}
    \centering
    \small
    \begin{tabularx}{0.48\textwidth}{l CCCC c} \toprule
      & \multicolumn{4}{c}{\textbf{Tag}} & \textbf{Form} \\
      \midrule
        & \textbf{P} & \textbf{R} & $\boldsymbol{F1}$ & \textbf{Acc}
        & \textbf{Acc} \\ \midrule
        Hebrew--BASE & \bf 89.04 & 40.12 & 55.32 & 86.88 & 83.63 \\
        Hebrew--LIN & 87.07 & 62.35 & 72.66 & 90.5 & \bf 86.75 \\
        Hebrew--NN & 87.18 & \bf 62.96 & \bf 73.12 & \bf 90.62 & 86.25 \\
        \midrule
        Spanish--BASE & \bf 96.97 & 51.45 & 67.23 & 90.21 & 86.32 \\
        Spanish--LIN & 92.74 & \bf 73.95 & \bf 82.29 & 93.79 & 89.52 \\
        Spanish--NN & 95.34 & 72.35 & 82.27 & \bf 93.91 & \bf 89.65 \\
         \bottomrule
    \end{tabularx}
    \caption{Tag-level precision, recall, $F_1$ score, and accuracy
      and form-level accuracy for the baselines (``{}--BASE'') and for
      our approach (``{}--LIN'' is the linear parameterization,
      ``{}--NN'' is the neural parameterization).}
    \label{tab:intrinsic}
\end{table}

\begin{figure*}
    \centering
    \small
    \begin{subfigure}{}
    \begin{tikzpicture}
        \begin{axis}[
            axis x line*=bottom, axis y line*=left,
            ybar=0pt,
            ymin=0,
            ylabel={Gender Bias},
            bar width=8pt,
            ymajorgrids,
            yminorgrids,
            symbolic x coords={Esp, Fra, Heb, Ita},
            enlarge x limits={value=0.25, auto},
            xtick=data,
            xticklabel style={text height=.7em},
            nodes near coords align={horizontal}, every node near coord/.append style={rotate=90},
            width=0.47\linewidth, height=15em,
            legend style={at={(0.83,1.1)},draw=none,fill=none,legend columns=-1},,
            ]
            \addplot coordinates {(Esp, 2.346048781310269 ) (Fra, 3.9480287697580128 ) (Heb, 3.984826103009676 ) (Ita, 5.973503041267395 )};
\addplot coordinates {(Esp, 0.44720202983576357 ) (Fra, 4.214745525960569 ) (Heb, 4.889504246962698 ) (Ita, 5.782071453730265 )};
\addplot coordinates {(Esp, 0.4730175986713424 ) (Fra, 2.348674393362469 ) (Heb, 1.8195368741687976 ) (Ita, 5.000627655982971 )};
            \legend{Original, Swap, MRF}
        \end{axis}
        \end{tikzpicture}
        \end{subfigure}
        \begin{subfigure}{}
        \begin{tikzpicture}
        \begin{axis}[
            axis x line*=bottom, axis y line*=left,
            ybar=0pt,
            ylabel={Grammaticality},
            bar width=8pt,
            ymajorgrids,
            yminorgrids,
            symbolic x coords={Esp, Fra, Heb, Ita},
            enlarge x limits={value=0.25, auto},
            xtick=data,
            xticklabel style={text height=.7em},
            nodes near coords align={horizontal}, every node near coord/.append style={rotate=90},
            width=0.47\linewidth, height=15em,
            legend style={at={(0.83,1.1)},draw=none,fill=none,legend columns=-1},
            ]
            \addplot coordinates {(Esp, 2.7331929860428033 ) (Fra, 1.6211015863551035 ) (Heb, 2.3897884406541525 ) (Ita, 1.5970661457379658 )};
\addplot coordinates {(Esp, 1.0149491376398152 ) (Fra, 1.6508987689459766 ) (Heb, 2.321306172170137 ) (Ita, 1.8520674800872803 )};
\addplot coordinates {(Esp, 2.115588519103739 ) (Fra, 1.6592156677334398 ) (Heb, 1.6244150952288978 ) (Ita, 1.723520939350128 )};
            \legend{Original, Swap, MRF}
        \end{axis}
        \end{tikzpicture}
        \end{subfigure}
        \caption{Gender stereotyping (left) and grammaticality (right) using the original corpus, the
  corpus following CDA using na\"{i}ve swapping of gendered words
  (``Swap''), and the corpus following CDA using our approach
  (``MRF'').\looseness=-1}
    \label{fig:bias}
\end{figure*}
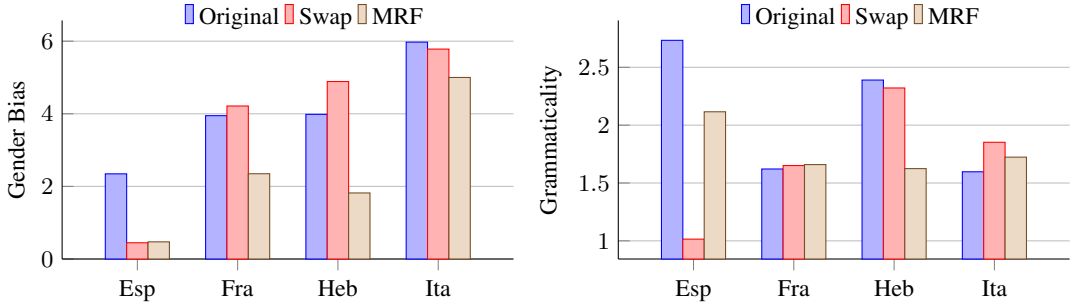

\paragraph{Results.}

We present the results in \cref{tab:intrinsic}. Recall is consistently
significantly lower than precision. As expected, the baselines have
the highest precision (though not by much). This is because they
reflect well-known rules for each language. That said, they have lower
recall than our approach because they fail to capture more subtle
relationships.

For both languages, our approach
struggles with conjunctions. For example, consider the phrase
\word{{\'e}l es un ingeniero y escritor} (\word{he is an engineer and
  a writer}). Replacing \word{ingeniero} with \word{ingeniera} does
not necessarily result in \word{escritor} being changed to
\word{escritora}. This is because two nouns do not normally need to
have the same gender when they are conjoined. Moreover, our MRF does
not include co-reference information, so it cannot tell that, in this
case, both nouns refer to the same person. We do not include
co-reference information in our MRF because this would create cycles
and inference would no longer be exact. Additionally, the lack of
co-reference information means that, for Spanish, our approach
fails to convert nouns that are noun-modifiers or indirect objects of
verbs.

Somewhat surprisingly, the neural parameterization does not outperform
the linear parameterization. We proposed the neural parameterization
to allow parameter sharing among edges with different parts of speech
and labels; however, this parameter sharing does not seem to make a
difference in practice, so the linear parameterization is sufficient.\looseness=-1

\subsection{Extrinsic Evaluation}\label{sec:extrinsic}

We extrinsically evaluate our approach by assessing the extent to
which it reduces gender stereotyping. Following
\newcite{DBLP:journals/corr/abs-1807-11714}, focus on neural language
models. We choose language models over word embeddings because
standard measures of gender stereotyping for word embeddings cannot~be
applied to morphologically rich languages.

As our measure of gender stereotyping, we compare the log ratio of the
prefix probabilities under a language model $P_{\textrm{lm}}$ for
gendered, animate nouns, such as \word{ingeniero}, combined with four
adjectives: \word{good}, \word{bad}, \word{smart}, and
\word{beautiful}. The translations we use for these adjectives are given in \cref{sec:translation}. We chose the first two adjectives because they
should be used equally to describe men and women, and the latter two
because we expect that they will reveal gender stereotypes. For
example, consider\looseness=-1
\begin{equation*}\label{eq:bias}
    \log  \frac{\sum_{\mathbf{x} \in \Sigma^*} \Plm(\textit{\textsc{bos} } \textit{El } \textit{ingeniero } \textit{bueno } \mathbf{x})}{\sum_{\mathbf{x} \in \Sigma^*}\Plm(\textit{\textsc{bos} } \textit{La } \textit{ingeniera } \textit{buena }\mathbf{x})}.
\end{equation*}
If this log ratio is close to 0, then the language model is as likely
to generate sentences that start with \word{el ingeniero bueno}
(\word{the good male engineer}) as it is to generate sentences that
start with \word{la ingeniera bueno} (\word{the good female
  engineer}).  If the log ratio is negative, then the language model
is more likely to generate the feminine form than the masculine form,
while the opposite is true if the log ratio is positive. In practice,
given the current gender disparity in engineering, we would expect the
log ratio to be positive. If, however, the language model were trained
on a corpus to which our CDA approach had been applied, we would then
expect the log ratio to be much closer to zero.

Because our approach is specifically intended to yield sentences that are grammatical, we additionally consider the following log ratio (i.e., the grammatical phrase over the ungrammatical phrase):
\begin{equation*}\label{eq:grammar}
    \log  \frac{\sum_{\mathbf{x} \in \Sigma^*} \Plm(\textit{\textsc{bos} } \textit{El } \textit{ingeniero } \textit{bueno } \mathbf{x})}{\sum_{\mathbf{x} \in \Sigma^*}\Plm(\textit{\textsc{bos} } \textit{El } \textit{ingeniera } \textit{bueno }\mathbf{x})}.
\end{equation*}

\begin{table}
    \centering
    \small
    \begin{tabularx}{0.48\textwidth}{rCC}
        \toprule
         \bf Language & \bf No. Animate Noun Pairs & \bf \% of Animate Sentences \\
         \midrule
         Hebrew & 95 & 20\% \\
         Spanish & 259 & 20\% \\
         Italian & 150 & 10\% \\
         French & 216 & 7\% \\
         \bottomrule
    \end{tabularx}
    \caption{Animate noun statistics.}
    \label{tab:anim}
\end{table}

We trained the linear parameterization using UD treebanks for Spanish,
Hebrew, French, and Italian (see \cref{tab:data}). For each of the
four languages, we parsed one million sentences from Wikipedia (May 2018 dump) using
\citet{DBLP:journals/corr/DozatM16}'s parser and extracted taggings
and lemmata using the method of \citet{D15-1272}. We automatically
extracted an animacy gazetteer from WordNet \cite{wordnet} and then
manually filtered the output for correctness. We provide the size of the languages' animacy gazetteers and the percentage of automatically parsed sentences
that contain an animate noun in \cref{tab:anim}. For each
sentence containing a noun in our animacy gazetteer, we created a copy of
the sentence, intervened on the noun, and then used our approach to
transform the sentence. For sentences containing more than one animate
noun, we generated a separate sentence for each possible combination
of genders. Choosing which sentences to duplicate is a
difficult task. For example, \word{alem\'{a}n} in Spanish can refer to
either a German man or the German language; however, we have no way of
distinguishing between these two meanings without additional
annotations. Multilingual animacy detection \cite{C18-1001} might help
with this challenge; co-reference
information might additionally help.\looseness=-1

For each language, we trained the BPE-RNNLM baseline open-vocabulary language model of
\citet{DBLP:journals/corr/abs-1804-08205} using the original corpus,
the corpus following CDA using na\"{i}ve swapping of gendered words,
and the corpus following CDA using our approach. We then computed
gender stereotyping and grammaticality as described above. We provide
example phrases in \cref{tab:lm}; we provide a more extensive list of phrases in
\cref{app:queries}.\looseness=-1

\paragraph{Results}

\begin{figure}
\small
\centering
\begin{tikzpicture}
        \begin{axis}[
            axis x line*=bottom, axis y line*=left,
            ybar=0pt,
            ylabel={Gender Bias},
            bar width=17pt,
            ymajorgrids,
            yminorgrids,
            minor y tick num=4,
            symbolic x coords={Original, Swap, MRF},
            enlarge x limits={value=0.25, auto},
            xtick=data,
            xticklabel style={text height=.7em},
            ytick={-5,0,5},
            nodes near coords align={horizontal}, every node near coord/.append style={rotate=90},
            width=\linewidth, height=16em,
            legend style={at={(1,1.05)},draw=none,fill=white},,
            ]
            \addplot coordinates {(Original,  -4.802478037382427 ) (Swap, -3.2850340040106523 ) (MRF, -2.1700485380072343 )};
            \addplot coordinates {(Original,  4.26326824016681 ) (Swap, 1.3179550957748507 ) (MRF, 0.965970638980948 )};
            \legend{Feminine, Masculine}
        \end{axis}
        \end{tikzpicture}
\caption{Gender stereotyping for words that are stereotyped toward men
  or women in Spanish using the original corpus, the
  corpus following CDA using na\"{i}ve swapping of gendered words
  (``Swap''), and the corpus following CDA using our approach
  (``MRF'').\looseness=-1}
        \label{fig:esp_bias}
\end{figure}
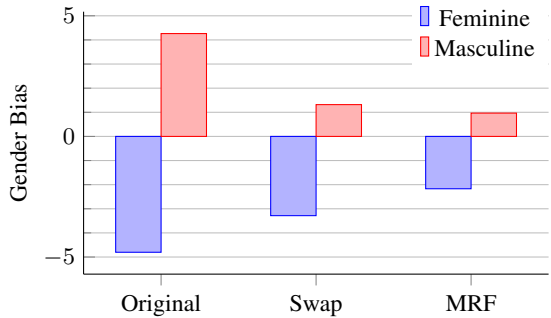

\cref{fig:bias} demonstrates depicts gender stereotyping and
grammaticality for each language using the original corpus, the corpus
following CDA using na\"{i}ve swapping of gendered words, and the
corpus following CDA using our approach. It is immediately apparent
that our approch reduces gender stereotyping. On average, our approach
reduces gender stereotyping by a factor of 2.5 (the lowest and highest
factors are 1.2 (Ita) and 5.0 (Esp), respectively). We expected that
na\"{i}ve swapping of gendered words would also reduce gender
stereotyping. Indeed, we see that this simple heuristic reduces gender
stereotyping for some but not all of the languages. For Spanish, we
also examine specific words that are stereotyped toward men or
women. We define a word to be stereotyped toward one gender if 75\% of
its occurrences are of that gender. \cref{fig:esp_bias} suggests a
clear reduction in gender stereotyping for specific words that are
stereotyped toward men or women.\looseness=-1

The grammaticality of the corpora following CDA differs between
languages. That said, with the exception of Hebrew, our approach
either sacrifices less grammaticality than na\"{i}ve swapping of
gendered words and sometimes increases grammaticality over the
original corpus. Given that we know the model did not perform as accurately for Hebrew (see \cref{tab:intrinsic}), this finding is not surprising.

\begin{table}
\centering
\small
\begin{tabularx}{0.48\textwidth}{lCCC}\toprule
\bf Phrase & \bf Original & \bf Swap & \bf MRF \\ \midrule
1. \word{El ingeniero bueno} &  -27.6 & -27.8 & -28.5 \\
2. \word{La ingeniera buena} & -31.3 & -31.6 & -30.5 \\
3. \word{*El ingeniera bueno} & -32.2 & -27.1 & -33.5 \\
4. \word{*La ingeniero buena} & -33.2 & -32.8 & -33.6 \\ \midrule
Gender stereotyping & 3.7 & 6.2 & 2 \\
Grammaticality & 3.25 & 0.25 & 4.05 \\ \bottomrule
\end{tabularx}
\caption{Prefix log-likelihoods of Spanish phrases using the original corpus, the corpus following CDA using na\"{i}ve swapping of gendered words (``Swap''), and the corpus following CDA using our approach (``MRF''). Phrases 1 and 2 are grammatical, while phrases 3 and 4 are not (dentoted by ``*'').
Gender stereotyping is measured using phrases 1 and 2. Grammaticality is measured using phrases 1 and 3 and using phrases 2 and 4; these scores are then averaged.\looseness=-1}
    \label{tab:lm}
\end{table}

\section{Related Work}

In contrast to previous work, we focus on mitigating gender
stereotypes in languages with rich morphology---specifically languages
that exhibit gender agreement. To date, the NLP community has focused
on approaches for detecting and mitigating gender stereotypes in
English. For example, \citet{bolukbasi2016} proposed a way of
mitigating gender stereotypes in word embeddings while preserving
meanings; \citet{DBLP:journals/corr/abs-1807-11714} studied gender
stereotypes in language models; and \citet{N18-2002} introduced a
novel Winograd schema for evaluating gender stereotypes in
co-reference resolution. The most closely related work is that of
\citet{zhao2018gender}, who used CDA to reduce gender stereotypes in
co-reference resolution; however, their approach yields ungrammatical
sentences in morphologically rich languages. Our approach is
specifically intended to yield grammatical sentences when applied to
such languages.
\citet{Habash:2019} also focused on morphologically rich languages, specifically Arabic, but in the context of gender identification in machine translation.

\section{Conclusion}

We presented a new approach for converting between masculine-inflected
and feminine-inflected noun phrases in morphologically rich languages. To
do this, we introduced a Markov random field with an optional neural
parameterization that infers the manner in which a sentence must
change to preserve morpho-syntactic agreement when altering the
grammatical gender of particular nouns. To the best of our knowledge,
this task has not been studied previously. As a result, there is no
existing annotated corpus of paired sentences that can be used as
``ground truth.'' Despite this limitation, we evaluated our approach
both intrinsically and extrinsically, achieving promising results. For
example, we demonstrated that our approach reduces gender stereotyping
in neural language models. Finally, we also identified avenues for
future work, such as the inclusion of co-reference information.

\section*{Acknowledgments}
R. Cotterell acknowledges a Facebook Fellowship.\looseness=-1

\bibliography{debias-cda-mrf}

\begin{thebibliography}{31}
\expandafter\ifx\csname natexlab\endcsname\relax\def\natexlab#1{#1}\fi

\bibitem[{Bolukbasi et~al.(2016)Bolukbasi, Chang, Zou, Saligrama, and
  Kalai}]{bolukbasi2016}
Tolga Bolukbasi, Kai{-}Wei Chang, James~Y. Zou, Venkatesh Saligrama, and
  Adam~Tauman Kalai. 2016.
\newblock \href
  {http://papers.nips.cc/paper/6228-man-is-to-computer-programmer-as-woman-is-to-homemaker-debiasing-word-embeddings}
  {Man is to computer programmer as woman is to homemaker? debiasing word
  embeddings}.
\newblock In \emph{Advances in Neural Information Processing Systems 29: Annual
  Conference on Neural Information Processing Systems 2016}, pages 4349--4357.

\bibitem[{Bond and Paik(2012)}]{wordnet}
Francis Bond and Kyonghee Paik. 2012.
\newblock A survey of {W}ord{N}ets and their licenses.
\newblock In \emph{Proceedings of the 6th Global WordNet Conference (GWC
  2012)}, Matsue.
\newblock 64--71.

\bibitem[{Coates(1987)}]{coates2015women}
Jennifer Coates. 1987.
\newblock \emph{Women, Men and Language: {A} Sociolinguistic Account of Sex
  Differences in Language}.
\newblock Longman.

\bibitem[{Corbett(1991)}]{corbett1991}
Greville~G. Corbett. 1991.
\newblock \emph{Gender}.
\newblock Cambridge University Press.

\bibitem[{Corbett(2012)}]{features}
Greville~G. Corbett. 2012.
\newblock \emph{Features}.
\newblock Cambridge University Press.

\bibitem[{Cotterell et~al.(2017)Cotterell, Kirov, Sylak-Glassman, Walther,
  Vylomova, Xia, Faruqui, K{\"u}bler, Yarowsky, Eisner, and
  Hulden}]{cotterell-etal-2017-conll}
Ryan Cotterell, Christo Kirov, John Sylak-Glassman, G{\'e}raldine Walther,
  Ekaterina Vylomova, Patrick Xia, Manaal Faruqui, Sandra K{\"u}bler, David
  Yarowsky, Jason Eisner, and Mans Hulden. 2017.
\newblock \href {https://doi.org/10.18653/v1/K17-2001} {{C}o{NLL}-{SIGMORPHON}
  2017 shared task: Universal morphological reinflection in 52 languages}.
\newblock In \emph{Proceedings of the {C}o{NLL} {SIGMORPHON} 2017 Shared Task:
  Universal Morphological Reinflection}, pages 1--30, Vancouver. Association
  for Computational Linguistics.

\bibitem[{Cotterell et~al.(2016)Cotterell, Kirov, Sylak-Glassman, Yarowsky,
  Eisner, and Hulden}]{cotterell2016}
Ryan Cotterell, Christo Kirov, John Sylak-Glassman, David Yarowsky, Jason
  Eisner, and Mans Hulden. 2016.
\newblock The {SIGMORPHON} 2016 shared task---morphological reinflection.
\newblock In \emph{Proceedings of the 2016 Meeting of SIGMORPHON}, Berlin,
  Germany. Association for Computational Linguistics.

\bibitem[{Crawford(2013)}]{crawford2013}
Kate Crawford. 2013.
\newblock \href {https://hbr.org/2013/04/the-hidden-biases-in-big-data} {The
  hidden biases in big data}.

\bibitem[{De{-}Arteaga et~al.(2019)De{-}Arteaga, Romanov, Wallach, Chayes,
  Borgs, Chouldechova, Geyik, Kenthapadi, and
  Kalai}]{DBLP:conf/fat/De-ArteagaRWCBC19}
Maria De{-}Arteaga, Alexey Romanov, Hanna~M. Wallach, Jennifer~T. Chayes,
  Christian Borgs, Alexandra Chouldechova, Sahin~Cem Geyik, Krishnaram
  Kenthapadi, and Adam~Tauman Kalai. 2019.
\newblock \href {https://doi.org/10.1145/3287560.3287572} {Bias in bios: {A}
  case study of semantic representation bias in a high-stakes setting}.
\newblock In \emph{Proceedings of the Conference on Fairness, Accountability,
  and Transparency, FAT* 2019, Atlanta, GA, USA, January 29-31, 2019}, pages
  120--128.

\bibitem[{Dixon et~al.(2018)Dixon, Li, Sorensen, Thain, and
  Vasserman}]{dixon2018}
Lucas Dixon, John Li, Jeffrey Sorensen, Nithum Thain, and Lucy Vasserman. 2018.
\newblock \href
  {www.aies-conference.com/wp-content/papers/main/AIES\_2018\_paper\_9. pdf}
  {Measuring and mitigating unintended bias in text classification}.

\bibitem[{Dozat and Manning(2016)}]{DBLP:journals/corr/DozatM16}
Timothy Dozat and Christopher~D. Manning. 2016.
\newblock \href {http://arxiv.org/abs/1611.01734} {Deep biaffine attention for
  neural dependency parsing}.
\newblock \emph{CoRR}, abs/1611.01734.

\bibitem[{Dryer and Haspelmath(2013)}]{wals}
Matthew~S. Dryer and Martin Haspelmath, editors. 2013.
\newblock \emph{WALS Online}.
\newblock Max Planck Institute for Evolutionary Anthropology, Leipzig.

\bibitem[{Garg et~al.(2017)Garg, Schiebinger, Jurafsky, and Zou}]{garg2018}
Nikhil Garg, Londa Schiebinger, Dan Jurafsky, and James Zou. 2017.
\newblock \href {http://arxiv.org/abs/1711.08412} {Word embeddings quantify 100
  years of gender and ethnic stereotypes}.
\newblock \emph{CoRR}, abs/1711.08412.

\bibitem[{Habash et~al.(2019)Habash, Bouamor, and Chung}]{Habash:2019}
Nizar Habash, Houda Bouamor, and Christine Chung. 2019.
\newblock Automatic gender identification and reinflection in arabic.
\newblock In \emph{Proceedings of the 1st ACL Workshop on Gender Bias for
  Natural Language Processing}, Florence, Italy.

\bibitem[{Jahan et~al.(2018)Jahan, Chauhan, and Finlayson}]{C18-1001}
Labiba Jahan, Geeticka Chauhan, and Mark Finlayson. 2018.
\newblock \href {http://aclweb.org/anthology/C18-1001} {A new approach to
  animacy detection}.
\newblock In \emph{Proceedings of the 27th International Conference on
  Computational Linguistics}, pages 1--12. Association for Computational
  Linguistics.

\bibitem[{Kingma and Ba(2014)}]{DBLP:journals/corr/KingmaB14}
Diederik~P. Kingma and Jimmy Ba. 2014.
\newblock \href {http://arxiv.org/abs/1412.6980} {Adam: {A} method for
  stochastic optimization}.
\newblock \emph{CoRR}, abs/1412.6980.

\bibitem[{Koller and Friedman(2009)}]{mrf}
Daphne Koller and Nir Friedman. 2009.
\newblock \emph{Probabilistic graphical models: {P}rinciples and techniques}.
\newblock MIT Press.

\bibitem[{Lu et~al.(2018)Lu, Mardziel, Wu, Amancharla, and
  Datta}]{DBLP:journals/corr/abs-1807-11714}
Kaiji Lu, Piotr Mardziel, Fangjing Wu, Preetam Amancharla, and Anupam Datta.
  2018.
\newblock \href {http://arxiv.org/abs/1807.11714} {Gender bias in neural
  natural language processing}.
\newblock \emph{CoRR}, abs/1807.11714.

\bibitem[{Mielke and Eisner(2018)}]{DBLP:journals/corr/abs-1804-08205}
Sabrina~J. Mielke and Jason Eisner. 2018.
\newblock \href {http://arxiv.org/abs/1804.08205} {Spell once, summon anywhere:
  {A} two-level open-vocabulary language model}.
\newblock \emph{CoRR}, abs/1804.08205.

\bibitem[{M{\"u}ller et~al.(2015)M{\"u}ller, Cotterell, Fraser, and
  Sch{\"u}tze}]{D15-1272}
Thomas M{\"u}ller, Ryan Cotterell, Alexander Fraser, and Hinrich Sch{\"u}tze.
  2015.
\newblock \href {https://doi.org/10.18653/v1/D15-1272} {Joint lemmatization and
  morphological tagging with lemming}.
\newblock In \emph{Proceedings of the 2015 Conference on Empirical Methods in
  Natural Language Processing}, pages 2268--2274. Association for Computational
  Linguistics.

\bibitem[{Nivre et~al.(2018)Nivre, Abrams, Agi{\'c}, Ahrenberg, Antonsen,
  Aplonova, Aranzabe, Arutie, Asahara, Ateyah, Attia, Atutxa, Augustinus,
  Badmaeva, Ballesteros, Banerjee, Bank, Barbu~Mititelu, Basmov, Bauer,
  Bellato, Bengoetxea, Berzak, Bhat, Bhat, Biagetti, Bick, Blokland, Bobicev,
  B{\"o}rstell, Bosco, Bouma, Bowman, Boyd, Burchardt, Candito, Caron, Caron,
  Cebiro{\u g}lu~Eryi{\u g}it, Cecchini, Celano, {\v C}{\'e}pl{\"o}, Cetin,
  Chalub, Choi, Cho, Chun, Cinkov{\'a}, Collomb, {\c C}{\"o}ltekin, Connor,
  Courtin, Davidson, de~Marneffe, de~Paiva, Diaz~de Ilarraza, Dickerson, Dirix,
  Dobrovoljc, Dozat, Droganova, Dwivedi, Eli, Elkahky, Ephrem, Erjavec,
  Etienne, Farkas, Fernandez~Alcalde, Foster, Freitas, Gajdo{\v s}ov{\'a},
  Galbraith, Garcia, G{\"a}rdenfors, Garza, Gerdes, Ginter, Goenaga, Gojenola,
  G{\"o}k{\i}rmak, Goldberg, G{\'o}mez~Guinovart, Gonz{\'a}les~Saavedra,
  Grioni, Gr{\=u}z{\={\i}}tis, Guillaume, Guillot-Barbance, Habash, Haji{\v c},
  Haji{\v c}~jr., H{\`a}~M{\~y}, Han, Harris, Haug, Hladk{\'a}, Hlav{\'a}{\v
  c}ov{\'a}, Hociung, Hohle, Hwang, Ion, Irimia, Ishola, Jel{\'{\i}}nek,
  Johannsen, J{\o}rgensen, Ka{\c s}{\i}kara, Kahane, Kanayama, Kanerva, Katz,
  Kayadelen, Kenney, Kettnerov{\'a}, Kirchner, Kopacewicz, Kotsyba, Krek, Kwak,
  Laippala, Lambertino, Lam, Lando, Larasati, Lavrentiev, Lee,
  L{\^e}~H{\`{\^o}}ng, Lenci, Lertpradit, Leung, Li, Li, Li, Lim, Ljube{\v
  s}i{\'c}, Loginova, Lyashevskaya, Lynn, Macketanz, Makazhanov, Mandl,
  Manning, Manurung, M{\u a}r{\u a}nduc, Mare{\v c}ek, Marheinecke,
  Mart{\'{\i}}nez~Alonso, Martins, Ma{\v s}ek, Matsumoto, {McDonald}, Mendon{\c
  c}a, Miekka, Misirpashayeva, Missil{\"a}, Mititelu, Miyao, Montemagni, More,
  Moreno~Romero, Mori, Mori, Mortensen, Moskalevskyi, Muischnek, Murawaki,
  M{\"u}{\"u}risep, Nainwani, Navarro~Hor{\~n}iacek, Nedoluzhko, Ne{\v
  s}pore-B{\=e}rzkalne, Nguy{\~{\^e}}n~Th{\d i}, Nguy{\~{\^e}}n Th{\d i}~Minh,
  Nikolaev, Nitisaroj, Nurmi, Ojala, Ol{\'u}{\`o}kun, Omura, Osenova,
  {\"O}stling, {\O}vrelid, Partanen, Pascual, Passarotti, Patejuk,
  Paulino-Passos, Peng, Perez, Perrier, Petrov, Piitulainen, Pitler, Plank,
  Poibeau, Popel, Pretkalni{\c n}a, Pr{\'e}vost, Prokopidis,
  Przepi{\'o}rkowski, Puolakainen, Pyysalo, R{\"a}{\"a}bis, Rademaker,
  Ramasamy, Rama, Ramisch, Ravishankar, Real, Reddy, Rehm, Rie{\ss}ler,
  Rinaldi, Rituma, Rocha, Romanenko, Rosa, Rovati, Roșca, Rudina, Rueter,
  Sadde, Sagot, Saleh, Samard{\v z}i{\'c}, Samson, Sanguinetti, Saul{\={\i}}te,
  Sawanakunanon, Schneider, Schuster, Seddah, Seeker, Seraji, Shen, Shimada,
  Shohibussirri, Sichinava, Silveira, Simi, Simionescu, Simk{\'o}, {\v
  S}imkov{\'a}, Simov, Smith, Soares-Bastos, Spadine, Stella, Straka,
  Strnadov{\'a}, Suhr, Sulubacak, Sz{\'a}nt{\'o}, Taji, Takahashi, Tanaka,
  Tellier, Trosterud, Trukhina, Tsarfaty, Tyers, Uematsu, Ure{\v s}ov{\'a},
  Uria, Uszkoreit, Vajjala, van Niekerk, van Noord, Varga, Villemonte de~la
  Clergerie, Vincze, Wallin, Wang, Washington, Williams, Wir{\'e}n,
  Woldemariam, Wong, Yan, Yavrumyan, Yu, {\v Z}abokrtsk{\'y}, Zeldes, Zeman,
  Zhang, and Zhu}]{ud}
Joakim Nivre, Mitchell Abrams, {\v Z}eljko Agi{\'c}, Lars Ahrenberg, Lene
  Antonsen, Katya Aplonova, Maria~Jesus Aranzabe, Gashaw Arutie, Masayuki
  Asahara, Luma Ateyah, Mohammed Attia, Aitziber Atutxa, Liesbeth Augustinus,
  Elena Badmaeva, Miguel Ballesteros, Esha Banerjee, Sebastian Bank, Verginica
  Barbu~Mititelu, Victoria Basmov, John Bauer, Sandra Bellato, Kepa Bengoetxea,
  Yevgeni Berzak, Irshad~Ahmad Bhat, Riyaz~Ahmad Bhat, Erica Biagetti, Eckhard
  Bick, Rogier Blokland, Victoria Bobicev, Carl B{\"o}rstell, Cristina Bosco,
  Gosse Bouma, Sam Bowman, Adriane Boyd, Aljoscha Burchardt, Marie Candito,
  Bernard Caron, Gauthier Caron, G{\"u}l{\c s}en Cebiro{\u g}lu~Eryi{\u g}it,
  Flavio~Massimiliano Cecchini, Giuseppe G.~A. Celano, Slavom{\'{\i}}r {\v
  C}{\'e}pl{\"o}, Savas Cetin, Fabricio Chalub, Jinho Choi, Yongseok Cho,
  Jayeol Chun, Silvie Cinkov{\'a}, Aur{\'e}lie Collomb, {\c C}a{\u g}r{\i} {\c
  C}{\"o}ltekin, Miriam Connor, Marine Courtin, Elizabeth Davidson,
  Marie-Catherine de~Marneffe, Valeria de~Paiva, Arantza Diaz~de Ilarraza,
  Carly Dickerson, Peter Dirix, Kaja Dobrovoljc, Timothy Dozat, Kira Droganova,
  Puneet Dwivedi, Marhaba Eli, Ali Elkahky, Binyam Ephrem, Toma{\v z} Erjavec,
  Aline Etienne, Rich{\'a}rd Farkas, Hector Fernandez~Alcalde, Jennifer Foster,
  Cl{\'a}udia Freitas, Katar{\'{\i}}na Gajdo{\v s}ov{\'a}, Daniel Galbraith,
  Marcos Garcia, Moa G{\"a}rdenfors, Sebastian Garza, Kim Gerdes, Filip Ginter,
  Iakes Goenaga, Koldo Gojenola, Memduh G{\"o}k{\i}rmak, Yoav Goldberg, Xavier
  G{\'o}mez~Guinovart, Berta Gonz{\'a}les~Saavedra, Matias Grioni, Normunds
  Gr{\=u}z{\={\i}}tis, Bruno Guillaume, C{\'e}line Guillot-Barbance, Nizar
  Habash, Jan Haji{\v c}, Jan Haji{\v c}~jr., Linh H{\`a}~M{\~y}, Na-Rae Han,
  Kim Harris, Dag Haug, Barbora Hladk{\'a}, Jaroslava Hlav{\'a}{\v c}ov{\'a},
  Florinel Hociung, Petter Hohle, Jena Hwang, Radu Ion, Elena Irimia, {\d
  O}l{\'a}j{\'{\i}}d{\'e} Ishola, Tom{\'a}{\v s} Jel{\'{\i}}nek, Anders
  Johannsen, Fredrik J{\o}rgensen, H{\"u}ner Ka{\c s}{\i}kara, Sylvain Kahane,
  Hiroshi Kanayama, Jenna Kanerva, Boris Katz, Tolga Kayadelen, Jessica Kenney,
  V{\'a}clava Kettnerov{\'a}, Jesse Kirchner, Kamil Kopacewicz, Natalia
  Kotsyba, Simon Krek, Sookyoung Kwak, Veronika Laippala, Lorenzo Lambertino,
  Lucia Lam, Tatiana Lando, Septina~Dian Larasati, Alexei Lavrentiev, John Lee,
  Phuong L{\^e}~H{\`{\^o}}ng, Alessandro Lenci, Saran Lertpradit, Herman Leung,
  Cheuk~Ying Li, Josie Li, Keying Li, {KyungTae} Lim, Nikola Ljube{\v s}i{\'c},
  Olga Loginova, Olga Lyashevskaya, Teresa Lynn, Vivien Macketanz, Aibek
  Makazhanov, Michael Mandl, Christopher Manning, Ruli Manurung, C{\u a}t{\u
  a}lina M{\u a}r{\u a}nduc, David Mare{\v c}ek, Katrin Marheinecke, H{\'e}ctor
  Mart{\'{\i}}nez~Alonso, Andr{\'e} Martins, Jan Ma{\v s}ek, Yuji Matsumoto,
  Ryan {McDonald}, Gustavo Mendon{\c c}a, Niko Miekka, Margarita
  Misirpashayeva, Anna Missil{\"a}, C{\u a}t{\u a}lin Mititelu, Yusuke Miyao,
  Simonetta Montemagni, Amir More, Laura Moreno~Romero, Keiko~Sophie Mori,
  Shinsuke Mori, Bjartur Mortensen, Bohdan Moskalevskyi, Kadri Muischnek, Yugo
  Murawaki, Kaili M{\"u}{\"u}risep, Pinkey Nainwani, Juan~Ignacio
  Navarro~Hor{\~n}iacek, Anna Nedoluzhko, Gunta Ne{\v s}pore-B{\=e}rzkalne,
  Luong Nguy{\~{\^e}}n~Th{\d i}, Huy{\`{\^e}}n Nguy{\~{\^e}}n Th{\d i}~Minh,
  Vitaly Nikolaev, Rattima Nitisaroj, Hanna Nurmi, Stina Ojala, Ad{\'e}day{\d
  o} Ol{\'u}{\`o}kun, Mai Omura, Petya Osenova, Robert {\"O}stling, Lilja
  {\O}vrelid, Niko Partanen, Elena Pascual, Marco Passarotti, Agnieszka
  Patejuk, Guilherme Paulino-Passos, Siyao Peng, Cenel-Augusto Perez, Guy
  Perrier, Slav Petrov, Jussi Piitulainen, Emily Pitler, Barbara Plank, Thierry
  Poibeau, Martin Popel, Lauma Pretkalni{\c n}a, Sophie Pr{\'e}vost, Prokopis
  Prokopidis, Adam Przepi{\'o}rkowski, Tiina Puolakainen, Sampo Pyysalo,
  Andriela R{\"a}{\"a}bis, Alexandre Rademaker, Loganathan Ramasamy, Taraka
  Rama, Carlos Ramisch, Vinit Ravishankar, Livy Real, Siva Reddy, Georg Rehm,
  Michael Rie{\ss}ler, Larissa Rinaldi, Laura Rituma, Luisa Rocha, Mykhailo
  Romanenko, Rudolf Rosa, Davide Rovati, Valentin Roșca, Olga Rudina, Jack
  Rueter, Shoval Sadde, Beno{\^{\i}}t Sagot, Shadi Saleh, Tanja Samard{\v
  z}i{\'c}, Stephanie Samson, Manuela Sanguinetti, Baiba Saul{\={\i}}te, Yanin
  Sawanakunanon, Nathan Schneider, Sebastian Schuster, Djam{\'e} Seddah,
  Wolfgang Seeker, Mojgan Seraji, Mo~Shen, Atsuko Shimada, Muh Shohibussirri,
  Dmitry Sichinava, Natalia Silveira, Maria Simi, Radu Simionescu, Katalin
  Simk{\'o}, M{\'a}ria {\v S}imkov{\'a}, Kiril Simov, Aaron Smith, Isabela
  Soares-Bastos, Carolyn Spadine, Antonio Stella, Milan Straka, Jana
  Strnadov{\'a}, Alane Suhr, Umut Sulubacak, Zsolt Sz{\'a}nt{\'o}, Dima Taji,
  Yuta Takahashi, Takaaki Tanaka, Isabelle Tellier, Trond Trosterud, Anna
  Trukhina, Reut Tsarfaty, Francis Tyers, Sumire Uematsu, Zde{\v n}ka Ure{\v
  s}ov{\'a}, Larraitz Uria, Hans Uszkoreit, Sowmya Vajjala, Daniel van Niekerk,
  Gertjan van Noord, Viktor Varga, Eric Villemonte de~la Clergerie, Veronika
  Vincze, Lars Wallin, Jing~Xian Wang, Jonathan~North Washington, Seyi
  Williams, Mats Wir{\'e}n, Tsegay Woldemariam, Tak-sum Wong, Chunxiao Yan,
  Marat~M. Yavrumyan, Zhuoran Yu, Zden{\v e}k {\v Z}abokrtsk{\'y}, Amir Zeldes,
  Daniel Zeman, Manying Zhang, and Hanzhi Zhu. 2018.
\newblock \href {http://hdl.handle.net/11234/1-2895} {Universal dependencies
  2.3}.
\newblock {LINDAT}/{CLARIN} digital library at the Institute of Formal and
  Applied Linguistics ({{\'U}FAL}), Faculty of Mathematics and Physics, Charles
  University.

\bibitem[{Pearl(1988)}]{pearl}
Judea Pearl. 1988.
\newblock \emph{Probabilistic reasoning in intelligent systems: {N}etworks of
  plausible inference}.
\newblock Morgan Kaufmann Publishers.

\bibitem[{Rabiner and Juang(1986)}]{rabiner1986introduction}
Lawrence~R. Rabiner and Biing-Hwang Juang. 1986.
\newblock An introduction to hidden {M}arkov models.
\newblock \emph{IEEE ASSP Magazine}, 3(1):4--16.

\bibitem[{Rall(1981)}]{DBLP:books/sp/Rall81}
Louis~B. Rall. 1981.
\newblock \href {https://doi.org/10.1007/3-540-10861-0} {\emph{Automatic
  {D}ifferentiation: {T}echniques and {A}pplications}}, volume 120 of
  \emph{Lecture Notes in Computer Science}.
\newblock Springer.

\bibitem[{Rudinger et~al.(2017)Rudinger, May, and Van~Durme}]{W17-1609}
Rachel Rudinger, Chandler May, and Benjamin Van~Durme. 2017.
\newblock \href {https://doi.org/10.18653/v1/W17-1609} {Social bias in elicited
  natural language inferences}.
\newblock In \emph{Proceedings of the First ACL Workshop on Ethics in Natural
  Language Processing}, pages 74--79. Association for Computational
  Linguistics.

\bibitem[{Rudinger et~al.(2018)Rudinger, Naradowsky, Leonard, and
  Van~Durme}]{N18-2002}
Rachel Rudinger, Jason Naradowsky, Brian Leonard, and Benjamin Van~Durme. 2018.
\newblock \href {https://doi.org/10.18653/v1/N18-2002} {Gender bias in
  coreference resolution}.
\newblock In \emph{Proceedings of the 2018 Conference of the North American
  Chapter of the Association for Computational Linguistics: Human Language
  Technologies, Volume 2 (Short Papers)}, pages 8--14. Association for
  Computational Linguistics.

\bibitem[{Sutton et~al.(2018)Sutton, Lansdall{-}Welfare, and
  Cristianini}]{sutton2018}
Adam Sutton, Thomas Lansdall{-}Welfare, and Nello Cristianini. 2018.
\newblock \href {http://arxiv.org/abs/1806.06301} {Biased embeddings from wild
  data: Measuring, understanding and removing}.
\newblock \emph{CoRR}, abs/1806.06301.

\bibitem[{Wu et~al.(2018)Wu, Shapiro, and Cotterell}]{D18-1473}
Shijie Wu, Pamela Shapiro, and Ryan Cotterell. 2018.
\newblock \href {http://aclweb.org/anthology/D18-1473} {Hard non-monotonic
  attention for character-level transduction}.
\newblock In \emph{Proceedings of the 2018 Conference on Empirical Methods in
  Natural Language Processing}, pages 4425--4438. Association for Computational
  Linguistics.

\bibitem[{Zhao et~al.(2019)Zhao, Wang, Yatskar, Cotterell, Ordonez, and
  Chang}]{zhao2019}
Jieyu Zhao, Tianlu Wang, Mark Yatskar, Ryan Cotterell, Vicente Ordonez, and
  Kai-Wei Chang. 2019.
\newblock \href {https://www.aclweb.org/anthology/N19-1064} {Gender bias in
  contextualized word embeddings}.
\newblock In \emph{Proceedings of the 2019 Conference of the North {A}merican
  Chapter of the Association for Computational Linguistics: Human Language
  Technologies, Volume 1 (Long and Short Papers)}, pages 629--634, Minneapolis,
  Minnesota. Association for Computational Linguistics.

\bibitem[{Zhao et~al.(2017)Zhao, Wang, Yatskar, Ordonez, and Chang}]{zhao2017}
Jieyu Zhao, Tianlu Wang, Mark Yatskar, Vicente Ordonez, and Kai-Wei Chang.
  2017.
\newblock \href {https://doi.org/10.18653/v1/D17-1323} {Men also like shopping:
  Reducing gender bias amplification using corpus-level constraints}.
\newblock pages 2979--2989.

\bibitem[{Zhao et~al.(2018)Zhao, Wang, Yatskar, Ordonez, and
  Chang}]{zhao2018gender}
Jieyu Zhao, Tianlu Wang, Mark Yatskar, Vicente Ordonez, and Kai-Wei Chang.
  2018.
\newblock \href {https://doi.org/10.18653/v1/N18-2003} {Gender bias in
  coreference resolution: Evaluation and debiasing methods}.
\newblock pages 15--20.

\end{thebibliography}
\bibliographystyle{acl_natbib}

\clearpage
\appendix
\section{Belief Propagation Update Equations}\label{sec:bp}
Our belief propagation update equations are
\begin{align}
\mu_{i\rightarrow f}(m) &=
\prod_{f' \in N(i)\setminus\{f\}}
\mu_{f'\rightarrow i}(m)
\label{eq:msg-var} \\
\mu_{f_i\rightarrow i}(m) &= \phif(m)\,\mu_{i\rightarrow f_i}(m)
\label{eq:msg-fac1}
\end{align}
\begin{align}
&\mu_{f_{ij}\rightarrow i}(m) = {}\label{eq:msg-fac2} \notag\\
&\sum_{m'\in M} \psif(m', m \mid p_i, p_j, \ell)\,\mu_{j\rightarrow f_{ij}}(m')
 \\
&\mu_{f_{ij}\rightarrow j}(m) ={} \label{eq:msg-fac3} \notag \\
&\sum_{m'\in M} \psif(m, m' \mid  p_i, p_j, \ell)\,\mu_{i\rightarrow f_{ij}}(m')
\end{align}
where $N(i)$ returns the set of neighbouring nodes of node $i$.
The belief at any node is given by
\begin{equation}\label{eq:belief}
\beta(v)= \prod_{f\in N(v)}\mu_{f\rightarrow v}(m).
\end{equation}

\section{Adjective Translations}\label{sec:translation}
\cref{tab:fem} and \cref{tab:masc} contain the feminine and masculine translations of the four adjectives that we used.\looseness=-1

\begin{table}[htb]
\centering
\small
\begin{tabularx}{0.48\textwidth}{XXXXX}\toprule
\bf Adjective & \bf French & \bf Hebrew & \bf Italian & \bf Spanish \\ \midrule
\word{good} & \word{bonne} & \begin{cjhebrew}.twbh\end{cjhebrew} & \word{buona} & \word{buena} \\ %  טובה
\word{bad} & \word{mauvaise} & \begin{cjhebrew}r`h\end{cjhebrew} & \word{cattiva} & \word{mala} \\ % רעה
\word{smart} & \word{intelligente} & \begin{cjhebrew}.hkmh\end{cjhebrew} & \word{intelligenti} & \word{inteligente} \\ % חכמה
\word{beautiful} & \word{belle} & \begin{cjhebrew}yph\end{cjhebrew} & \word{bella} & \word{hermosa} \\ % יפה
\bottomrule
\end{tabularx}
\caption{Feminine translations of \word{good}, \word{bad}, \word{smart}, \word{beautiful} in French, Hebrew, Italian, and Spanish}
    \label{tab:fem}
\end{table}

\begin{table}[htb]
\centering
\small
\begin{tabularx}{0.48\textwidth}{XXXXX}\toprule
\bf Adjective & \bf French & \bf Hebrew & \bf Italian & \bf Spanish \\ \midrule
\word{good} & \word{bon} & \begin{cjhebrew}.twb\end{cjhebrew} & \word{buono} & \word{bueno} \\ % טוב
\word{bad} & \word{mauvais} & \begin{cjhebrew}r`\end{cjhebrew} & \word{cattivo} & \word{malo} \\ % רע
\word{smart} & \word{intelligent} & \begin{cjhebrew}.hkM\end{cjhebrew} & \word{intelligente} & \word{inteligente} \\ % חכם
\word{beautiful} & \word{bel} & \begin{cjhebrew}yph\end{cjhebrew} & \word{bello} & \word{hermoso} \\ % יפה
\bottomrule
\end{tabularx}
\caption{Masculine translations of \word{good}, \word{bad}, \word{smart}, \word{beautiful} in French, Hebrew, Italian, and Spanish}
    \label{tab:masc}
\end{table}
\newpage
\section{Extrinsic Evaluation Example Phrases}\label{app:queries}

For each noun in our animacy gazetteer, we generated sixteen phrases.
Consider the noun \word{engineer} as an example. We created four
phrases---one for each translation of \word{The good engineer},
\word{The bad engineer}, \word{The smart engineer}, and \word{The
  beautiful engineer}. These phrases, as well as their prefix
log-likelihoods are provided below in \cref{tab:query}.

\begin{table}[htb]
\centering
\small
\begin{tabularx}{0.48\textwidth}{lCCC}\toprule
\bf Phrase & \bf Original & \bf Swap & \bf MRF \\ \midrule
\word{El ingeniero bueno} & -27.63 & -27.80 & -28.50 \\
\word{La ingeniera buena} & -31.34 & -31.65 & -30.46 \\
\word{*El ingeniera bueno} & -32.22 & -27.06 & -33.49 \\
\word{*La ingeniero buena} & -33.22 & -32.80 & -33.56 \\
\word{El ingeniero mal} & -30.45 & -30.90 & -30.86 \\
\word{La ingeniera mala} & -31.03 & -29.63 & -30.59 \\
\word{*El ingeniera mal} & -34.19 & -30.17 & -35.15 \\
\word{*La ingeniero mala} & -33.09 & -30.80 & -33.81 \\
\word{El ingeniero inteligente} & -26.19 & -25.49 & -26.64 \\
\word{La ingeniera inteligente} & -29.14 & -26.31 & -27.57 \\
\word{*El ingeniera inteligente} & -29.80 & -24.99 & -30.77 \\
\word{*La ingeniero inteligente} & -31.00 & -27.12 & -30.16 \\
\word{El ingeniero hermoso} & -28.74 & -28.65 & -29.13 \\
\word{La ingeniera hermosa} & -31.21 & -29.25 & -30.04 \\
\word{*El ingeniera hermoso} & -32.54 & -27.97 & -33.83 \\
\word{*La ingeniero hermosa} & -33.55 & -30.35 & -32.96 \\
\bottomrule
\end{tabularx}
\caption{Prefix log-likelihoods of Spanish phrases using the original corpus, the corpus following CDA using na\"{i}ve swapping of gendered words (``Swap''), and the corpus following CDA using our approach (``MRF''). Ungrammatical phrases are denoted by ``*''.}
    \label{tab:query}
\end{table}

\end{document}